\documentclass[journal]{IEEEtran}

\usepackage{lineno,hyperref}
\modulolinenumbers[5]

\usepackage[table]{xcolor}
\usepackage{array}
\newcolumntype{L}[1]{>{\raggedright\let\newline\\\arraybackslash\hspace{0pt}}m{#1}}
\newcolumntype{C}[1]{>{\centering\let\newline\\\arraybackslash\hspace{0pt}}m{#1}}
\newcolumntype{R}[1]{>{\raggedleft\let\newline\\\arraybackslash\hspace{0pt}}m{#1}}
\usepackage{amsmath}
\usepackage{subfig}
\usepackage{graphicx}
\usepackage{url}
\usepackage{textcomp}

\usepackage[ruled,linesnumbered,vlined]{algorithm2e}

\graphicspath{{Figures/}}
\DeclareGraphicsExtensions{.pdf,.png,.jpg}

\usepackage{amssymb}
\usepackage[inline]{enumitem}
\setlist{nolistsep} 
\setlist[itemize]{leftmargin=*} 
\usepackage{booktabs} 
\usepackage{multirow} 
\usepackage{siunitx}  
\usepackage{calc}  
\usepackage{todonotes}
\usepackage{color}

\usepackage[normalem]{ulem}
\usepackage{xcolor}
\makeatletter
\def\squiggly{\bgroup \markoverwith{\textcolor{red}{\lower3.5\p@\hbox{\sixly \char58}}}\ULon}
\font\sixly=lasyb10 scaled 652
\makeatother

\newcommand{\newtext}[1]{\textcolor{black}{#1}}

\def \figurename {Fig.}
\def \figurenamelong {Figure}
\newcommand{\figref}[1]{\figurename~\ref{#1}}
\newcommand{\Figref}[1]{\figurenamelong~\ref{#1}}

\def \sectionname {Section}
\newcommand{\secref}[1]{\sectionname~\ref{#1}}

\def \algoname {Algorithm}
\newcommand{\algoref}[1]{\algoname~\ref{#1}}

\def \eg {e.g., }
\def \ie {i.e., }

\def \vs {vs.\ }
\def \ea {\textit{et al.\ }}

\DeclareMathOperator*{\argmin}{argmin}

\def \questionname {Q}
\newcommand{\qref}[1]{\questionname\ref{#1}}
\newcounter{qcounter}
\newenvironment{question}[1]{\refstepcounter{qcounter}\textbf{(\questionname\theqcounter)} #1}{}

\def\tslot{\Delta t^\text{slot}}
\def\tdepart{\Delta t^\text{depart}}
\def\tcharge{\Delta t^\text{charge}}
\def\tflex{\Delta t^\text{flex}}

\def\Smax{S_\text{max}}
\def\Hmax{H_\text{max}}
\def\Nmax{N_\text{max}}
\def\Cbau{C_\text{BAU}}
\def\Copt{C_\text{opt}}
\def\CRL{C_\text{RL}}

\begin{document}

\title{Definition and evaluation of model-free coordination of electrical vehicle charging with reinforcement learning}

\author{Nasrin~Sadeghianpourhamami,~\IEEEmembership{Student~Member,~IEEE}, Johannes Deleu,
        and~Chris~Develder,~\IEEEmembership{Senior~Member,~IEEE}
\thanks{N. Sadeghianpourhamami is with IDLab, Dept.\
	of Information Technology, Ghent University, Ghent,
	Belgium, e-mail: \href{mailto:nasrin.sadeghianpourhamami@ugent.be}{nasrin.sadeghianpourhamami@ugent.be}}
\thanks{J. Deleu is with IDLab, Dept.\
	of Information Technology, Ghent University, Ghent,
	Belgium, e-mail: \href{mailto:Johannes.deleu@ugent.be}{johannes.deleu@ugent.be}}
\thanks{C. Develder is with IDLab, Dept.\
of Information Technology, Ghent University - imec, Ghent,
Belgium, e-mail: \href{mailto:chris.develder@ugent.be}{chris.develder@ugent.be} (see \url{http://users.ugent.be/\~cdvelder/}).}
}

\markboth{\today}%
{N.~Sadeghianpourhamami \MakeLowercase{\textit{et al.}}: \textbf{Definition and evaluation of model-free coordination of electrical vehicle charging with reinforcement learning}}
\maketitle

\begin{abstract}
With the envisioned growth in deployment of electric vehicles (EVs), managing the joint load from EV charging stations through demand response (DR) approaches becomes more critical. Initial DR studies mainly adopt model predictive control and thus require accurate models of the control problem (\eg a customer behavior model), which are to a large extent uncertain for the EV scenario. Hence, model-free approaches, especially based on reinforcement learning (RL) are an attractive alternative. In this paper, we propose a new Markov decision process (MDP) formulation in the RL framework, to jointly coordinate a set of EV charging stations. State-of-the-art algorithms either focus on a single EV, or perform the control of an aggregate of EVs in multiple steps (\eg aggregate load decisions in one step, then a step translating the aggregate decision to individual connected EVs). On the contrary, we propose an RL approach to jointly control the whole set of EVs at once. 
We contribute a new MDP formulation, with a scalable state representation that is independent of the number of EV charging stations. Further, we use a batch reinforcement learning algorithm, \ie an instance of fitted Q-iteration, to learn the optimal charging policy. We analyze its performance using simulation experiments based on a real-world EV charging data.
More specifically, we 
\begin{enumerate*}[label=(\roman*)]
\item explore the various settings in training the RL policy (\eg duration of the period with training data),
\item compare its performance to an oracle all-knowing benchmark (which provides an upper bound for performance, relying on information that is not available or at least imperfect in practice),
\item analyze performance over time, over the course of a full year to evaluate possible performance fluctuations (e.g, across different seasons), and
\item demonstrate the generalization capacity of a learned control policy to larger sets of charging stations.
\end{enumerate*}
\end{abstract}


\IEEEpeerreviewmaketitle

\section*{Nomenclature}
\begin{IEEEdescription}[\IEEEusemathlabelsep\IEEEsetlabelwidth{$C(s,\textbf{u}_s,s')$}]
	\item[$s$] State
	\item[$s'$] The next state from $s$
	\item[$\tdepart$] Time left until departure
	\item[$\tcharge$] Time needed for charging completion
	\item[$\tflex$] Flexibility (time charging can be delayed)
	\item[$N_s$] Number of connected EVs in state $s$
	\item[$\mathcal{V}_t$] Set of EVs in the system at time $t$
	\item[$\textbf{x}_s$] Aggregate demand in state $s$
	\item[$t$] Timeslot 
	\item[$\tslot$] Duration of a decision slot
	\item[$\Smax$] Maximum number of decision slots
	\item[$\Hmax$] Maximum connection time
	\item[$\Nmax$] Number of charging stations jointly being coordinated
	\item[$\textbf{u}_s$] Action taken in state $s$
	\item[$\textbf{U}_s$] Set of possible actions from state $s$
	\item[$\textbf{x}_s^\text{total}(d)$] Total number of EVs on the $d$\textsuperscript{th} upper diagonal of $\textbf{x}_s$
	\item[$C^\text{demand}$] Cost of total power consumption
	\item[$C^\text{penalty}$] Penalty cost for unfinished charging
	\item[$C(s,\textbf{u}_s,s')$] Instantaneous cost of state transition
	\item[$\mathcal{B}^\text{test}$] Test set
	\item[$\mathcal{B}^\text{train}$] Training set 
	\item[$\Delta t$] Training data time span
	\item[$C_{\pi}$] Normalized cost of policy $\pi$
	\item[$\Cbau$] Normalized cost of business-as-usual policy
	\item[$\CRL$] Normalized cost of the learned policy
	\item[$\Copt$] Normalized cost of optimum solution \end{IEEEdescription}


\section{Introduction}

\IEEEPARstart{D}{emand} response (DR) algorithms aim to coordinate the energy consumption of customers in a smart grid to ensure demand-supply balance and 
reliable network performance.
In initial DR studies, the demand response problem usually is cast as a model predictive control (MPC) approach (\eg \cite{Afram2014, Ma2012}), typically formulated as an optimization problem to minimize the customer's electricity bill or maximize the energy provider's profit, subject to various operating constraints (\eg physical characteristics of the devices, customer preferences, distributed energy resource constraints and energy market constraints).
However, the widespread deployment of such model-based DR algorithms in the smart grid is limited for the following reasons:
\begin{enumerate*}[label=(\roman*)]
\item heterogeneity of the end user loads, difference in user behavioral patterns and uncertainty surrounding their behavior makes the modeling task very challenging \cite{Sadeghianpourhamami2016};
\item model-based DR algorithms are difficult to transfer from one scenario to the other, since the model designed for one group of users or applications is likely to require customization/tweaking for application to different groups.
\end{enumerate*}

Recently, reinforcement learning (RL) has emerged to facilitate model-free control for coordinating the user flexibility in DR algorithms. In RL-based approaches, the DR problem is defined in the form of a Markov decision process (MDP). A coordinating agent interacts with the environment (\ie DR participating customers, energy providers, energy market prices, etc.) and takes control actions while aiming to maximize the long term expected reward (or minimize the long term expected cost). In other words, the agent learns by taking actions and observing the outcomes (\ie states) and the rewards/costs in an iterative process. The DR objective (\eg load flattening, load balancing) is achieved by appropriately designing the reward/cost signal. Hence, reinforcement learning based approaches do not need an explicit model of user flexibility behavior or the energy pricing information a priori. This facilitates more practical and generally applicable DR schemes compared to model-based approaches.  

One of the main challenges of RL-based DR approaches is the curse of dimensionality due to the continuity and scale of the state and the action spaces: this hinders the applicability of RL-based DR for large-scale problems. In this paper, we focus on formulating a scalable RL-based DR algorithm to coordinate the charging of a group of electric vehicle (EV) charging stations, which generalizes to various group sizes and EV charging rates. In fact, current literature only offers a limited amount of model-free solutions for jointly coordinating the charging of multiple EV charging stations, as surveyed briefly in \secref{sec:RL:related}.

Such existing RL-based DR solutions are either developed for an individual EV or need a heuristic (which does not guarantee an optimum solution) to obtain the aggregate load of multiple EV charging stations during the learning process. Indeed, a scalable Markov decision process (MDP) formulation that generalizes to a collection of EV charging stations with different characteristics (\eg charging rates, size) does not exist in current literature. In this paper we take the first step to fill this gap by proposing an MDP and explore its performance in simulation experiments. Note that the model we present is a further refinement of our initially proposed state and action representation listed in \cite{sadeghianpourhamami2018eenergyB} (which did not consider sizable experimental results yet, and merely proposed a first MDP formulation).
More precisely, in this paper:
\begin{itemize}
	\item We define a new MDP with compact state and action space representations, in the sense that they do \emph{not} linearly scale with the number of EV charging stations (thus EVs), they can generalize to collections of various sizes and they can be extended to cope with heterogeneous charging rates (see \secref{sec:RL:MDP}), 
	\item We adopt batch reinforcement learning (fitted Q-iteration \cite{Riedmiller2005}) with function approximation to find the best EV charging policy (see \secref{sec:RL:BRL}),
	\item We quantitatively explore the performance of the proposed reinforcement learning approach, through simulations using real-world data to run experiments covering 10 and 50 charging stations (using the setup detailed in \secref{sec:RL:experiment}), answering the following research questions (see \secref{sec:RL:results}):
	\begin{itemize}[label=]
		\item \begin{question}\label{q:RL:training}\label{q:RL:first}What are appropriate parameter settings\footnote{The parameters of interest are \begin{enumerate*}[label=(\roman*)]
        \item time span of the training data, and 
        \item number of sampled trajectories from the decision trees. 
        \end{enumerate*} For details see \secref{sec:RL:size} and \secref{sec:RL:setting}.} of the input training data?\end{question} 
		\item \begin{question}\label{q:RL:optimal}How does the RL policy perform compare to an optimal all-knowing oracle algorithm?\end{question}
		\item \begin{question}\label{q:RL:time-variance}How does that performance vary over time (\ie from one month to the next) using realistic data?\end{question}
		\item \begin{question}\label{q:RL:last}\label{q:RL:generalization}Does a learned approach generalize to different EV group sizes?\end{question}
	\end{itemize}
\end{itemize}
We summarize our conclusions and list open issues to be addressed in future work in \secref{sec:RL:conclusion}.

\section{Related Work}
\label{sec:RL:related}

With growing EV adoption, also the amount of available (and realistic) EV data increased. Hence, data-driven approaches to coordinate EV charging gained attention, with reinforcement learning (RL) as a notable example. 
For example, Shi \ea\cite{Shi2011} adopt an RL-based approach and phrase an MDP to learn to control the charging and discharging of an \emph{individual EV} under price uncertainty for providing vehicle-to-grid (V2G) services.
Their MDP has 
\begin{enumerate*}[label=(\roman*)]
\item a state space based on the hourly electricity price, state-of-charge and time left till departure),
\item an action space to decide between charging (either to fulfill the demand or provide frequency regulation), delaying the charging and discharging for frequency regulation\footnote{Frequency regulation is a so-called ancillary service for the power grid, and entails actions to keep the frequency of the alternating current grid within tight bounds, by instantaneous adjustments to balance generation and demand.}, and
\item unknown state transition probabilities.
\end{enumerate*}
The reward is defined as the energy payment of charging and discharging or the capacity payment (for the provided frequency regulation service).
%
Chis \ea\cite{Chis2017} use batch RL to learn the charging policy of again an \emph{individual EV}, to reduce the long-term electricity costs for the EV owner. An MDP framework is used to represent this problem, where
\begin{enumerate*}[label=(\roman*)]
\item the state space consists of timing variables, minimum charging price for a current day and price fluctuation between the current and the next day, while
\item the action is the amount of energy to consume in a day.
\end{enumerate*}
Cost savings of 10\%-50\% are reported for simulations using real-world pricing data.
Opposed to these cost-minimizing approaches assuming time-varying prices, as a first case study for our joint control of a group of EV charging stations, we will focus first on a load flattening scenario (\ie electricity prices are assumed constant, but peaks need to be avoided).

\newtext{In contrast to \cite{Shi2011} and \cite{Chis2017}, which consider the charging of a single EV, Claessens \ea\cite{ Claessens2013} use batch RL to learn a collective charging plan for a \emph{group of EVs} in the optimization step of their previously proposed three step DR approach  \cite{Vandael2013}. Their three step DR approach constitutes an aggregation step, an optimization step, and a real-time control step. In the aggregation step, individual EV constraints are aggregated. In the optimization step, the aggregated constraints are used by the batch RL agent to learn the collective charging policy for the EV fleet, which is translated to a sequence of actions (\ie aggregated power consumption values for each decision slot) to minimize energy supply costs. Finally, in the real-time control step a priority based heuristic algorithm is used dispatch the energy corresponding to the action determined in the optimization step from the individual EVs.
Vandael \ea\cite{Vandael2015} also use batch RL to learn a cost-effective day-ahead consumption plan for a \emph{group of EVs}.}
Their formulation has two decision phases, 
\begin{enumerate*}[label=(\roman*)]
\item day-ahead and
\item intra-day.
\end{enumerate*}
In the first decision phase, the aggregator predicts the energy required for charging its EVs for the next day, and purchases this amount in the day-ahead market. This first decision phase is modeled as an MDP.
In the second decision phase, the aggregator communicates with the EVs to control their charging, based on the amount of energy purchased in the day-ahead market. The amount of the energy to be consumed by each connected EV is calculated using a heuristic priority-based algorithm and is communicated to the respective EV. The local decision making process by each EV is modeled using an MDP where the state space is represented by the charged energy of the EV, the action space is defined by charging power and the reward function is based on the deviations from the requested charging power. The fitted Q-iteration (FQI) algorithm is used to obtain the best policy.

\newtext{
Note that our work is different from \cite{Claessens2013} and \cite{Vandael2015} in two aspects:
\begin{enumerate*}[label=(\roman*)]
\item unlike \cite{Claessens2013} and \cite{Vandael2015}, our proposed approach does not take the control decisions in separate steps (\ie taking aggregate energy consumption in one step and coordinating individual EV charging in a second step to meet the already decided energy consumption) and instead it takes decisions directly and jointly for all individual EVs using an efficient representation of an aggregate state of a group of EVs, hence
\item our approach does not need a heuristic algorithm, but instead learns the aggregate load while finding an optimum policy to flatten the load curve. 
\end{enumerate*}}
We now describe our MDP model, and subsequently the batch reinforcement learning approach to train it.

\section{Markov Decision Process}
\label{sec:RL:MDP}
The high-level goal of the proposed EV charging approach is to minimize the long term cost of charging a group of EVs for an aggregator in a real-time decision-making scenario. In this paper, we focus on the scenario of load flattening (\ie more advanced DR objectives are left for future work): we aim to minimize the peak-to-average ratio of the aggregate load curve of a group of EVs. Technically, we adopt a convex cost function that sums the squares of the total consumption over all timeslots within the decision time horizon. We regard this problem as a sequential decision making problem and formulate it using an MDP with unknown transition probabilities. 

\begin{algorithm}[!t]
	\caption{Binning algorithm for creating the aggregate state representation.}
	\label{alg:RL:binning}
	\newcommand\mycommfont[1]{\small\ttfamily{#1}}
	\SetCommentSty{mycommfont}
	\SetKwInOut{Input}{Input}
	\SetKwInOut{Output}{Output}

	\Input{
		$\mathcal{V}_t = \{(\tdepart_1,\tcharge_1), \ldots, (\tdepart_{N_s},\tcharge_{N_s})\}$
		\\
	}
	\Output{
		Aggregate state $\textbf{x}_s$, matrix of size $\Smax \times \Smax$
	}
	Initialize $\textbf{x}_s$ with zeros\\
	\ForEach{$n = 1, \ldots, N_s$}{ \label{alg-line:RL:begincount}
		\tcp{count number of EVs in each $(i,j)$ bin}
		$i = \left \lceil{\frac{\tdepart_n}{\tslot}}\right \rceil$ \label{alg-line:RL:i}\\
		$j = \left \lceil{\frac{\tcharge_n}{\tslot}}\right \rceil$ \label{alg-line:RL:j}\\
		$\textbf{x}_{s}(i,j) \leftarrow \textbf{x}_{s}(i,j)+1$\label{alg-line:RL:up}\label{alg-line:RL:endcount}
	}
	\KwRet $\textbf{x}_s/N_\textit{max}$ \label{alg-line:RL:normalize}
	
\end{algorithm}

\subsection{State Space}
An EV charging session is characterized by:
\begin{enumerate*}[label=(\roman*)]
\item EV arrival time,
\item time left till departure ($\tdepart$),
\item requested energy and 
\item EV charging rate.
\end{enumerate*}
We translate the requested energy to time needed to complete the charging ($\tcharge$), implicitly assuming the same charging rates for all the EVs in a group.
Thus, if we have $N_s$ electric vehicles in the system, the (remaining times of) their sessions are represented as a set 
\begin{equation*}
\mathcal{V}_t = \{(\tdepart_1,\tcharge_1), \ldots, (\tdepart_{N_s},\tcharge_{N_s})\}.
\end{equation*}
Note that we do not assume a priori knowledge of future arrivals, and hence do not include the arrival time to characterize the (present) EVs. 

Each state $s$ is represented using two variables: timeslot (\ie $t \in \{1,\dots,\Smax\}$) and the aggregate demand (\ie $\textbf{x}_s$), hence $s=(t,\textbf{x}_s)$. \newtext{Inspired by \cite{Claessens2018},} aggregate demand at each given timeslot is \newtext{obtained via a binning algorithm (\ie \algoref{alg:RL:binning})} and is represented using a 2D grid, thus a matrix, with one axis representing $\tdepart$, the other $\tcharge$. As time progresses, cars will move towards lower $\Delta t^{\textit{depart}}$ cells, and (if charged) lower $\tcharge$ and $\tdepart$.\footnote{An extension to consider the variable charging rate is possible by binning the EVs in a 3D grid with charging rate as the third dimension.}
Given that time-of-day is likely to influence the expected evolution of the state $\textbf{x}_s$ (and hence the required response action we should take), we do include the timeslot $t$ as explicit part of the state.
%

Formally, the process to convert the set of sessions $\mathcal{V}_t$ (associated with EVs connected at a given time $t$) to the matrix $\textbf{x}_s$ is given by \algoref{alg:RL:binning}. The size of the matrix, $\Smax \times \Smax$ depends on the maximal connection time $\Hmax$, \ie the longest duration of an EV being connected to a charging station: $\Smax \triangleq \Hmax/\tslot$.
%

\begin{figure}[!t]
	\centering
	\includegraphics[scale=0.6]{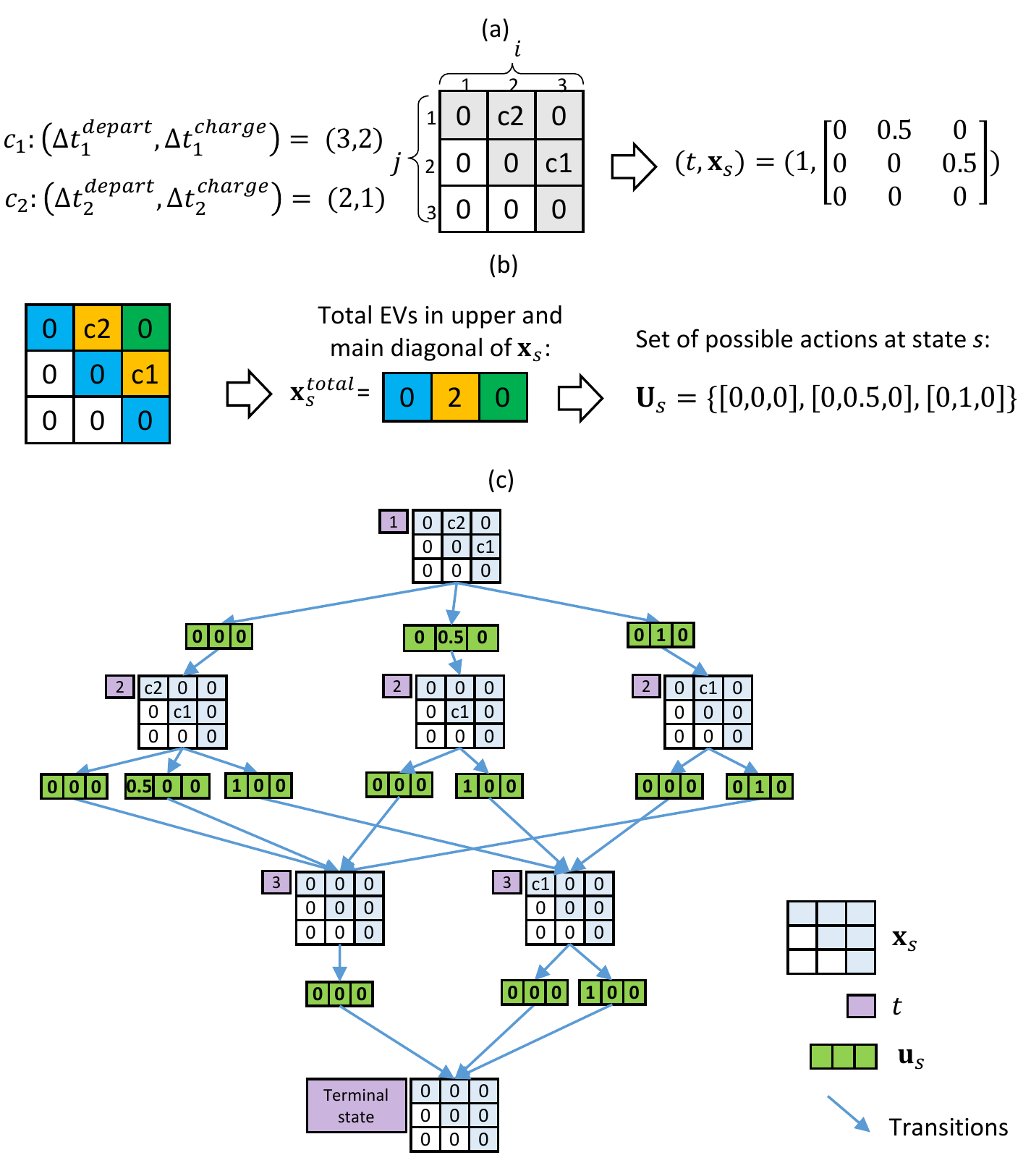}
	\caption{A simple example for $\Nmax = 2$ charging stations: (a)~state representation, (b)~possible action states, (c)~full decision tree over the horizon of $\Smax = 3$ slots.}
	\label{fig:RL:example1}
\end{figure}  

Each row/column of $\textbf{x}_s$ represents equidistant bins with edges on $\{0,\tslot,2\cdot\tslot,\ldots,\Smax\cdot\tslot\}$ and each matrix element in $\textbf{x}_s$ represents the number of EVs binned into it.
$\textbf{x}_s$ is initialized with zeros at the beginning of \algoref{alg:RL:binning}. Lines~\ref{alg-line:RL:begincount}--\ref{alg-line:RL:endcount} count the EVs with $\tdepart$ and $\tcharge$ values of the corresponding $(i,j)$-cell in the matrix. Finally, $\textbf{x}_s$ is normalized by $\Nmax$ (Line~\ref{alg-line:RL:normalize}). This normalization makes the state representation scale-free, \ie independent of the absolute group size $\Nmax$, thus aiming to generalize the formulated MDP (and the learned control policy) to a differently sized group of EV charging stations. 

For illustrative purposes, in \figref{fig:RL:example1} we sketch a simple scenario of $\Nmax = 2$ charging stations with a horizon of $\Smax = 3$~slots. 
Let us assume that at time $t = 1$ we have $N_s=2$ connecting cars: $\mathcal{V}_1 = \{(\Delta t^\text{{depart}}_1,\Delta t^{\text{charge}}_1) = (3,2), (\Delta t^\text{{depart}}_2,\Delta t^{\text{charge}}_2) = (2,1)\}$, with no other arrivals during the control horizon.
\Figref{fig:RL:example1} illustrates the resulting state space using the binning algorithm at the first time slot. The EVs are binned according to their $\tdepart$ and $\tcharge$ to a 2D grid of size $3\times3$. The resulting matrix is normalized by $\Nmax$ ($=2$ in this example). The shaded cells in the 2D grid of \figref{fig:RL:example1} indicate bins with $\Delta t^\text{{charge}}\leq \Delta t^{\text{depart}}$. EVs in these bins have enough time to complete their charging.

Note that $\textbf{x}_s$ not only summarizes the aggregated demand of connecting EVs (in terms of $\tdepart$ and $\tcharge$), but also the flexibility in terms of how long the charging can be delayed at state $s$ (denoted as $\tflex = \tdepart - \tcharge$) is inferred from the diagonals of $\textbf{x}_s$ using
\begin{equation}\label{eq:RL:flex}
\tflex(i,j) = j-i \qquad \forall i,j \in \{1,\dots, S_{\textit{max}}\}
\end{equation}
Equation~\eqref{eq:RL:flex} indicates that EVs binned into cells on the main diagonal of $\textbf{x}_s$ (\ie $i=j$) have zero flexibility while the ones binned into cells on the upper diagonals of $\textbf{x}_s$ are flexible charging requests. Negative $\tflex$, corresponding to lower diagonals of $\textbf{x}_s$ 
(\ie the white cells in the 2D grids of \figref{fig:RL:example1}),
indicates EVs for which the requested charging demand cannot be fulfilled.
In our formulation, we will ensure that EVs charging demands are never violated, using a penalty term in our cost function (see \secref{sec:RL:cost}). 

Finally, the size of $\textbf{x}_s$ and hence the size of the state $s$ is independent of $\Nmax$ and is only influenced by $\Smax$, thus $\Hmax$ and $\tslot$. This ensures scalability of the state representation to various group sizes of EV charging stations: the maximal number of cars $\Nmax$ does not impact the state size.

\subsection{Action Space}
The action to take in state $s$ is a decision whether (or not) to charge the connecting EVs with same $\tflex$ in the $\textbf{x}_s$ matrix. Such EVs are binned into the cells on the same diagonal of $\textbf{x}_s$ as explained in the previous section.
We indicate each diagonal of $\textbf{x}_s$ as $\textbf{x}_s(d)$ with $d = 0,\dots,\Smax-1$ where $\textbf{x}_s(0)$ is the main diagonal, $\textbf{x}_s(d)$ is the upper $d^{th}$ diagonal, and $\textbf{x}_s(-d)$ is the lower $d^{th}$ diagonal of $\textbf{x}_s$. We denote $\textbf{x}_s^\text{total}(d)$ as the total number of EVs in the cells on the $d^{th}$ diagonal. 

An action taken in state $s$ is defined as a vector $\textbf{u}_s$ of length $\Smax$.
For each individual car, we take a discrete action, \ie we either charge it at full power or not at all for the next timeslot.
This results in the element $d$ of the action vector $\textbf{u}_s$ being a number between 0 and 1: it amounts to charging the fraction of EVs in the corresponding $d^\textit{th}$ diagonal of $\textbf{x}_s$. 
The set of possible actions from state $s$ is denoted as $\textbf{U}_s$.


\Figref{fig:RL:example1}(b) illustrates how  $\textbf{U}_s$ is constructed at state $s$ using a color-coded representation of matrix $\textbf{x}_s$ and the corresponding vector $\textbf{x}_s^\text{total}$.
Note that we define the action vector for charging/delaying the cars on the main and upper diagonals of $\textbf{x}_s$ only 
(colored cells in the 2D grids representing $\textbf{x}_s$ in \figref{fig:RL:example1})
. This is a design choice to keep the action space relatively small and therefore easier to explore. In the next section, we define our cost function such that the EV charging is always completed before departure: no cars will end up in any of the lower diagonals, \ie the white cells in the 2D state grid of the figures.


\subsection{Cost function}
\label{sec:RL:cost}

The goal we envision in this paper is to flatten the aggregate charging load of a group of EVs while ensuring each EV's charging is completed before departure.\footnote{We assume only feasible requests are presented to the system, \ie $\tcharge \leq \tdepart$ for each EV.} Hence, our cost function associated with each state transition $(s,\textbf{u}_s,s')$ has two parts:
\begin{enumerate}[label=(\arabic*)]
	\item $ C^{\text{demand}}(\textbf{x}_s,\textbf{u}_s)$: the cost of the total power consumption from all the connected EVs for a decision slot, and 
	\item $C^{\text{penaltiy}}(\textbf{x}_{s'})$: the penalty for unfinished charging.
\end{enumerate} 

To achieve the load flattening objective, we choose the $C^{\text{demand}}$ to be a quadratic function of the total power consumption for a decision slot. The total power consumption for a decision slot is proportional to the number of EVs being charged, since 
we assume the same charging rate for all the EVs in a group. Hence, the first term of the cost function at state $s=(t,\textbf{x}_{s})$ is defined as
\begin{equation}\label{eq:RL:cost_demand}
C^{\text{demand}}(\textbf{x}_s,\textbf{u}_s) =\left( \sum_{d=0}^{S_\textit{max}-1}\textbf{x}_s^\text{total}(d) \> \textbf{u}_s(d)\right)^2
\end{equation}


The second term of the cost function is a penalty proportional to the unfinished charging in the next state $s'=(t_{s'},\textbf{x}_{s'})$ due to taking action $\textbf{u}_{s}$ in $s=(t,\textbf{x}_{s})$ and is defined as
\begin{equation}\label{eq:RL:penalty}
C^{\text{penalty}}(\textbf{x}_{s'}) = M\sum_{d=1}^{S_\textit{max}-1}\textbf{x}_{s'}^\text{total}(-d)
\end{equation}
%
The summation in \eqref{eq:RL:penalty} counts the number of EVs whose charging request is impossible to complete (EVs with $\tdepart_n < \Delta t^\textit{{charge}}_n$), \ie the cars that end up in the lower diagonals of the state matrix, in the next state $s' = (t+1,\textbf{x}_{s'})$ as a consequence of taking action $\textbf{u}_s$ at state $s = (t,\textbf{x}_s)$. $M$ is a constant penalty factor, which we set to be greater than $2\>\Nmax$ to ensure that any EV's charging is always completed before departing (\ie one incomplete EV is costlier than charging all EVs simultaneously).
Summing \eqref{eq:RL:cost_demand} and \eqref{eq:RL:penalty}, the total cost associated with each state transition $(s,\textbf{u}_s,s')$ is:
\begin{equation}\label{eq:RL:cost} 
C(s,\textbf{u}_s,s') = \left( \sum_{d=0}^{S_\textit{max}-1}\textbf{x}_s^\text{total}(d) \> \textbf{u}_s(d)\right)^2 + M \sum_{d=0}^{S_\textit{max}-1}\textbf{x}_{s'}^\text{total}(-d) 
\end{equation}

Note that in Equation~\eqref{eq:RL:cost} the cost is independent of the timeslot variable of the state space (\ie $t$) and depends only on the aggregate demand variable of the state (\ie $\textbf{x}_s$).
Indeed, the cost of a demand to be is set as a quadratic function of the total consumption to achieve the load flattening objective, and is time-independent. 
Still, we include the time component in the definition of the state to ensure that our formulations can easily be extended to other objectives (\eg reducing the cost under the time-of-use or pricing schemes).
Also, we use the time component for the function approximator of Algorithm~\ref{alg:RL:FQI} (see further, \secref{sec:RL:setting}).


\subsection{System Dynamics}

In the MDP framework, system dynamics (via the environment) are defined using transition probabilities
$P(s'|s, \textbf{u}_s)$. The transition probabilities from one state $s$ to the next $s'$ are unknown in the EV group charging problem due to the stochasticity of the EV arrivals and their charging demands. 
Perfect knowledge of EV arrivals and their charging demands during the control horizon would translate the problem into a decision tree depicted in \figref{fig:RL:example1}(c), where the cost of taking each action can be determined recursively using dynamic programming.
However, in absence of such knowledge, the transition probabilities need to be estimated through interactions with the environment by taking actions and observing the instantaneous cost of the resulting state transitions. The next section explains this approach.

\subsection{Learning Objective: State-Action Value Function}
\label{sec:RL:learningObj}
Note that $C(s,\textbf{u}_s, s')$ is the instantaneous cost an aggregator incurs when action $\textbf{u}_s$ is taken at state $s = (t,\textbf{x}_s)$ and leads to state $s'=(t+1,\textbf{x}_{s'})$. The objective is to find an optimum control policy $\pi^*: \textbf{S} \to \textbf{U}$ that minimizes the expected $T$-step return for any state in $\textbf{S}$. The expected $T$-step return starting from state $s=1$ and following a policy $\pi$ (\ie $\textbf{u}_s = \pi(s)$) is defined as:

\begin{equation}\label{eq:RL:Ereturn}
J^{\pi}_T(1) =\mathbb{E} \left[\sum_{s=1}^{T}{C(s,\textbf{u}_s,s')}\right]
\end{equation}

The policy $\pi$ is commonly characterized using a state-action value function (or Q-function):

\begin{equation}\label{eq:RL:Qfunc}
Q^{\pi}(s,\textbf{u}_s) =\mathbb{E} \left[C(s,\textbf{u}_s,s')+J^{\pi}_T(s')\right]
\end{equation}
where $Q^{\pi}(s,\textbf{u}_s)$ is cumulative return starting from state $s$, taking action $\textbf{u}_s$, and following policy $\pi$ afterwards. The optimal $Q^{\pi}(s,\textbf{u}_s)$, denoted as $Q^{*}(s,\textbf{u}_s)$, corresponds to:
\begin{equation}\label{eq:RL:Qopt}
Q^{*}(s,\textbf{u}_s) =\min_\pi Q^{\pi}(s,\textbf{u}_s)
\end{equation}

The $Q^{*}(s,\textbf{u}_s)$ satisfies the Bellman equation:
\begin{equation}\label{eq:RL:bellman}
Q^*(s,\textbf{u}_s) =\min_{\textbf{u} \in \textbf{U}} \mathbb{E}\left[C(s,\textbf{u}_s, s') + Q^*(s',\textbf{u})\right] 
\end{equation}

However, solving \eqref{eq:RL:bellman} requires the knowledge of the transition probabilities --- defining how the system moves from one state $s$ to the next $s'$ --- which are unknown in our setting. Hence, a learning algorithm should be used to obtain approximation $\widehat{Q}^*(s,\textbf{u})$. This can then be used to take control action $\textbf{u}_{s}$, following:
\begin{equation}\label{eq:RL:action}
\textbf{u}_{s} \in \argmin_{u \in \textbf{U}_s} \widehat{Q}^*(s,\textbf{u})
\end{equation}

\begin{algorithm}[!t]
	\caption{ Fitted Q-iteration using function approximation for estimating the $T$-step return}
	\label{alg:RL:FQI}
	\newcommand\mycommfont[1]{\small\ttfamily{#1}}
	\SetCommentSty{mycommfont}
	\SetKwInOut{Input}{Input}
	\SetKwInOut{Output}{Output}

	\Input{
		$\mathcal{F} = \{(s,\textbf{u}_s,s',C(s,\textbf{u}_s,s'))| s = 1, \ldots, |\mathcal{F}|\}$\; 
	}
	Initialize $\widehat{Q}_0$ to be zero everywhere on $\textbf{X} \times \textbf{U}$\; \label{alg-line:RL:initZeroes}
	\SetKwFor{For}{repeat}{times}{end repeat}
	\ForEach{$N = 1, \ldots, T$}{ \label{alg-line:RL:startItr}
		\ForEach{$(s, \textbf{u}_s, s', C(s,\textbf{u}_s,s'))) \in \mathcal{F}$}{\label{alg-line:RL:calQ}
			$Q_N(s,\textbf{u}_s) \leftarrow C(s,\textbf{u}_s,s') + \min\limits_{\textbf{u} \in \textbf{U}} \widehat{Q}_{N-1}(s',\textbf{u}) $
		}
		Use function approximator to obtain $\widehat{Q}_N$ from $\mathcal{T}_{\textit{reg}} = \big\{ ((s,\textbf{u}_s),Q_{N,s})|s = 1, \ldots, |\mathcal{F}| \big\} $ \label{alg-line:RL:functionApprox}
	}
	\KwRet $\widehat{Q}_T $
	
\end{algorithm}

\section{Batch Reinforcement Learning}
\label{sec:RL:BRL}

We adopt batch mode RL algorithms to approximate $\widehat{Q}^*(s,\textbf{u})$ from past experience instead of online interactions with the environment. In the batch mode RL approach, data collection is decoupled from the optimization. In other words, we use the historical EV data (\ie arrival/departures and energy demands) and a random policy to collect the experiences (\ie the state transitions and the associate costs) in form of $(s,\textbf{u}_s,s',C(s,\textbf{u}_s,s'))$ tuples. We use Fitted Q-iteration to approximate $\widehat{Q}^*(s,\textbf{u})$ from the collected tuples, detailed next.


\subsection{Fitted Q-iteration}
\label{sec:RL:FQI}

Fitted Q-iteration (FQI) is a batch mode RL algorithm, listed in \algoref{alg:RL:FQI}.
As input, FQI takes a set of past experiences, $\mathcal{F}$, in the form of tuples $(s,\textbf{u}_s,s',C(s,\textbf{u}_s,s'))$ where $C(s,\textbf{u}_s,s')$ is the immediate cost of a transition and in our case is calculated using Eq.~\eqref{eq:RL:cost}. The tuples are used to iteratively estimate the optimum action value function.

The state-action value function $Q$ is initialized with zeros on the state-action space (Line~\ref{alg-line:RL:initZeroes}) hence, $Q_1 = C(s,\textbf{u}_s,s')$ in the first iteration. In subsequent iterations, $Q_{N}$ is calculated for each tuple in $\mathcal{F}$ using the latest approximation of action-value function ($Q_{N-1}$) from the previous iteration (Line~4) to form a labeled dataset $\mathcal{T}_\textit{reg}$. This dataset is then used for regression, \ie by function approximation we estimate $Q_{N}$ for all possible state-action pairs (Line~\ref{alg-line:RL:functionApprox}). 

We will adopt a fully connected artificial neural network (ANN) as our function approximation. Further details on the ANN architecture used in our experiments are given in \secref{sec:RL:ANN}.

\subsection{The size of state-action space}
\label{sec:RL:size}
The input to FQI (\ie set $\mathcal{F}$) is constructed from past interactions with the environment (\ie randomly or deterministically taking actions from the action space of state $s=(t,\textbf{x}_s)$ and recording the tuple $(s,\textbf{u}_s,s',C(s,\textbf{u}_s,s'))$). The number of all possible actions from a given state $s$ is given by
\begin{equation}\label{eq:RL:growth}
\left|\textbf{U}_s\right| = \prod_{d=1}^{S_{\textit{max}}}\left(\textbf{x}_s^\textit{total}(d) + 1 \right)
\end{equation}

since for each flexibility $\tflex = d$ we can choose to charge between $[0, x^\text{total}_s(d)]$ cars.

The goal of the RL algorithm (hence the goal of the FQI) is to estimate the $T$-step return for every possible action from every possible state in the environment. Estimating the $T$-step return starting from a state $s$ leads to exploring a tree with an exponentially growing number of branches at the next steps. Hence, while the state and action representations are independent of the group size ($\Nmax$), the state-action space still grows exponentially with a growth rate given in Eq.~\eqref{eq:RL:growth}.
\newtext{Let us consider a charging lot of capacity $\Nmax=50$ and control horizon with $\Smax=10$.
In a state where 
all EV charging stations are occupied ($N_s = \Nmax = 50$),
there are at least 51 possible actions from that state, corresponding to a scenario where all the EVs have similar flexibility, hence located on the same diagonal of the state matrix (\ie $\textbf{x}_s^\text{total})=[50,0,0,0,0,0,0,0,0,0]$. For a state with $\textbf{x}_s^\text{total})=[5,5,5,5,5,5,5,5,5,5]$, there will be $\left|\textbf{U}_s\right|= (6)^{10}$ possible actions from that state only.} This indicates that it is not feasible to include the entire state-action space in set $\mathcal{F}$ as the input to the FQI and only a subset of the state-action space is provided.
We will therefore randomly sample trajectories from the decision tree with a branching factor of $|\textbf{U}_s|$.
This leads to the research question \textbf{\qref{q:RL:training}} (which is answered in \secref{sec:RL:results_learning}): How many sample trajectories from the state-action space are sufficient to learn an optimum policy for charging a real-world group of EVs with various group sizes?

%
%

\section{Experiment setup}
\label{sec:RL:experiment}
In this section, we outline the implementation details of the proposed RL-based DR approach.
\subsection{Data Preparation}
\label{sec:RL:dataset}
We base our analysis on real-world EV charging session transactions collected by ElaadNL since 2011 from 2500+ public charging stations deployed across Netherlands, as described and analyzed in \cite{Nasrin2017}. For each of the over 2M charging sessions (still growing), a transaction records the charging station ID, arrival time, departure time, requested energy and charging rate during the session. The EVs in this dataset are privately owned cars and thus comprise a mixture of different and a priori unknown car types. 

To represent the EV transactions in ElaadNL as state transitions $(s,\textbf{u}_s,s',C(s,\textbf{u}_s,s') )$, we first need to choose a reasonable size for the state and the action representations.
We set the maximum connection duration to $H_{\textit{max}}=24\,\textrm{h}$, since more than 98\% of the EV transactions in the ElaadNL dataset cover sessions of less than 24 hours \cite{Nasrin2017}.

We further set the duration of 
a decision timeslot, \ie the time granularity of control actions (\ie $\tslot = 2\,\textrm{h}$), resulting in $\Smax = \Hmax/\tslot = 12$. Hence, a state $s$ is represented by a scalar variable $t$ and a matrix $\textbf{x}_s$ of size $\Smax \times \Smax = 12 \times 12$. The corresponding action $\textbf{u}_s$ taken from state $s$ is a vector of length $12$ (with $1$ decision for each of the upper diagonals, one per flexibility window $\tflex$).
The motivation of choosing $\tslot=2\,\textrm{h}$ is to limit the branching factor $|\textbf{U}_s|$ (which depends on $\Smax$ in Eq.~\eqref{eq:RL:growth}) at each state, thus yielding a reasonable the state-action space size and allowing model training (specifically, the $\min$ operation in Line~4 of \algoref{alg:RL:FQI}) in a reasonable amount of time given our computation resources.\footnote{We use an Intel Xeon E5645 processor, 2.4\,GHz, 290\,GB RAM.}

Furthermore, we make the ElaadNL dataset episodic by assuming that all the EVs leave the charging stations before the end of a day, thus yielding an empty car park in between two consecutive days.\footnote{The charging demands of EVs are adjusted to ensure the requested charging can be fulfilled within $24$ hours.} 
We define such an episodic `day' to start at 7\,am and end 24\,h later (the day after at 7\,am).
The empty system state in between two episodes is always reached after $\Smax +1$ timeslots and is represented with aggregate demand matrix $\textbf{x}_s$ of all zeros. 
This ensures that while each day has a different starting state (depending on the arrivals in the first control slot and their energy demand), traversing the decision tree always leads to a unique terminal state (see \figref{fig:RL:example1}(c) for an exemplary decision tree).
This is motivated by Riedmiller \cite{Riedmiller2012}, who shows that, when learning with FQI and adopting a neural network as function approximator, having a terminal goal state stabilizes the learning process. It
ensures that all trajectories end up in a state where no further action/transition is possible and hence is characterized by an action-value of zero.  


To create a group of $\Nmax$ EV charging stations, we select the busiest $\Nmax$ charging stations (based on the number of recorded transactions in each station). For the analysis in this paper, 
we use two different subsets, one with the top-10, the other with the top-50 most busiest stations.                                                                 

\subsection{Algorithm Settings}
\label{sec:RL:setting}

Since $S_{max}=12$ in our settings, fitted Q-iteration (FQI) needs to estimate the 12-step return and we thus have 12 iterations in \algoref{alg:RL:FQI} .

\subsubsection{Creating set \texorpdfstring{$\mathcal{F}$}{F}}
\label{sec:RL:setF}

To create set $\mathcal{F}$, we begin from the starting state of a day characterized by $(t_1,\textbf{x}_1)$ and randomly choose an action from the set of possible actions in each state and observe the next state and the associated state transition cost until the terminal state\footnote{Recall that we consider an episodic setting, \ie case where the system empties (definitely after $\Smax$ timeslots).} is reached (\ie $(t_T,\textbf{x}_T)$).
The state transitions in each trajectory are recorded in the form of a tuple $(s,\textbf{u}_s,s',C(s,\textbf{u}_s,s'))$ in set $\mathcal{F}$. For our experiments, we randomly sample more than a single trajectory from each day to analyze the effect of the number of sampled trajectories on the performance of the proposed approach. The notion of a \textit{sample} in the following thus refers to a full trajectory from initial to terminal state of a day. 

\subsubsection{Neural network architecture}
\label{sec:RL:ANN}

We use an artificial neural network (ANN) that consists of an input layer, 2 hidden layers with ReLU activation function and an output layer. There are 128 and 64 neurons in the first and second hidden layers respectively. Since the ANN is used for linear regression, the output layer has a single neuron and a linear activation function.
Each state-action pair is fed to the input layer in form of a vector of length $\Smax^2 + \Smax + 1$, by reshaping the state $(t, s_t)$ and concatenating it with the action vector $\textbf{u}_s$ (of size $\Smax=12$). Recall that the state representation has a scalar time variable $t$ and an aggregate demand matrix $\textbf{x}_s$ of size $\Smax \times \Smax$, thus reshaped to a vector of size $\Smax^2+1$.
In our settings each state $s$ thus is represented as a vector of length 145 and each action $u_s$ as a vector of length 12. Inspired by Mnih \ea \cite{Mnih2015}, we also found that using Huber loss \cite{Huber1964} instead of mean-squared-error stabilizes the learning in our algorithm.


\subsection{Performance Evaluation Measure}
\label{sec:RL:measures}

To evaluate the performance of the proposed approach, we take ElaadNL transactions of 2015 and select the last 3 months as the test set, \ie
$\mathcal{B}^{\text{test}}=\{e_i|i = 274,\dots,365\}$ containing $|\mathcal{B}^{\text{test}}| = 92$ days.

We consider training sets of varying lengths (to determine the impact of training set size, see research question \qref{q:RL:training}), with $\Delta t \in \{1, 3, 5, 7, 9\}$ months.
For a given $\Delta t$ (\ie training data time span), we randomly pick 5 contiguous periods within the range of Jan.\ 1, 2015 until Sep.\ 30, 2015 (except for the case $\Delta t = 9$ months, since that covers the whole training data range). 
We define the training set for time span $\Delta t$ and run $j$ as $\mathcal{B}^{\text{train}}_{\Delta t,j}=\{e_i|i=e^{\text{start}}_{\Delta t,j},\dots,e^{\text{start}}_{\Delta t,j}+\Delta t -1\}$, where $e^{\text{start}}_{\Delta t,j}$ is the randomly selected starting date of the training set.

To evaluate the performance of the learned policy, we define the metric of \textit{normalized cost} relative to the cost of the optimum policy.
For each $\Delta t$ and $j$ we define it as
\begin{equation}\label{eq:RL:measure}
C_{\pi(\Delta t_j)}=\frac{1}{|\mathcal{B}^{\text{test}}|}\sum_{e\in \mathcal{B}^{\text{test}}}\frac{C_{\pi(\Delta t_j)}^e}{C^e_{\text{opt}}},
\end{equation} 
where $\pi(\Delta t_j)$ is a policy learned from the training data time span of $\Delta t$ at run $j$. Further, $C_{\pi(\Delta t_j)}^e$ is the cost of day $e$ under policy $\pi(\Delta t_j)$ and $C^e_{\text{opt}}$ is the cost of day $e$ using optimization (obtained from formulating the load flattening problem as a quadratic optimization problem).
A cost of a day $e$ under policy $\pi$ is calculated by summing the instantaneous cost (defined by Eq.~\eqref{eq:RL:cost}) of state transitions encountered when taking action according to the policy being evaluated (using Eq.~\eqref{eq:RL:action}). 

Clearly, if a learned policy achieves the optimum policy, then $C_{\pi(\Delta t_j)} = 1$. Further, we compare the performance of the learned policy not only with the optimum policy but also with the business-as-usual (BAU) policy where the charging of an EV starts immediately upon arrival. In the next section, we present our analysis using the normalized cost of BAU, optimum and learned policies denoted as $\Cbau$, $\Copt$ and $\CRL$ respectively.

\section{Experimental results}
\label{sec:RL:results}
In this section, we present experiments answering the aforementioned research questions \qref{q:RL:first}--\qref{q:RL:last}.
More specifically, we first evaluate the performance of the RL-based approach in coordinating the charging demand of $\Nmax =$ 10 and 50 charging stations as a function of training data time span and number of randomly sampled trajectories per day (\qref{q:RL:training}), comparing it to an uncontrolled business-as-usual scenario but also to the optimum strategy (\qref{q:RL:optimal}).
We then investigate how well the method works across various seasons, \ie whether performance varies strongly throughout the whole year (\qref{q:RL:time-variance}).
Finally, we check the scalability by training an agent on a group of $\Nmax =$ 10 EV charging stations and testing it on upscaled group sizes $\Nmax$ (\qref{q:RL:generalization}).


\begin{figure*}[!t]
	\centering
	\includegraphics[scale=0.7]{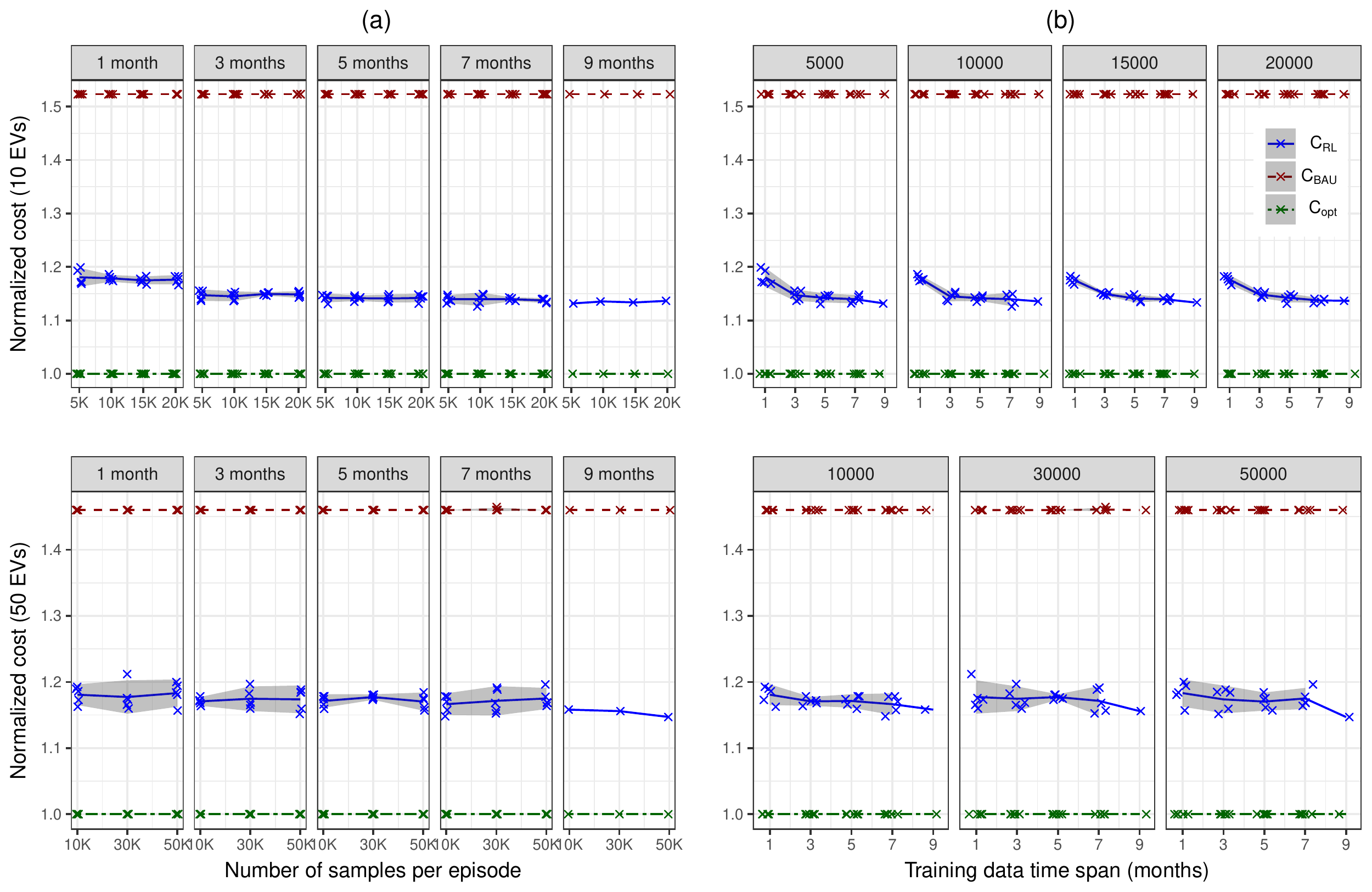}
	\caption{Normalized costs of learned policy ($\CRL$), BAU policy ($\Cbau$) and optimum solution ($\Copt$) for coordinating the charging of 10 (top row) and 50 (bottom row) EV charging stations, (a) normalized costs as a function of number of samples per day for various $\Delta t$s, and (b) normalized costs as a function of $\Delta t$ for various numbers of sample trajectories per training day.}
	\label{fig:RL:costJointEV}
\end{figure*}

\subsection{Learning the Charging Coordination (\qref{q:RL:training}--\qref{q:RL:optimal})}
\label{sec:RL:results_learning}

To answer \textbf{\qref{q:RL:first}} (\ie what are appropriate training data time span and number of sampled trajectories from the decision trees?), we study how the performance of the proposed RL approach varies in function of
\begin{enumerate*}[label=(\roman*)]
\item the time span covered by the training data (\ie $\Delta t$), and 
\item the number of sample trajectories per day of training data.
\end{enumerate*}
%
\Figref{fig:RL:costJointEV} compares the normalized cost of a learned policy with that of a BAU and optimum policy for varying $\Delta t$ and number of samples per training day, for the case of $\Nmax =$ 10 and 50 charging stations respectively.

\textit{Influence of the time span covered by the training data}:
\figref{fig:RL:costJointEV}(b) 
shows that
increasing $\Delta t$ from 1 month to 3 months and beyond reduces the normalized cost of the learned policy for both 10 and 50 charging stations.
Additionally, the performance gain when increasing $\Delta t$ from 1 to 3 months is bigger than for increasing $\Delta t$ beyond 3 months. This suggests that the RL approach needs at least 3 months of training data to reach maximal performance (in case of ElaadNL). 

\textit{Influence of the number of sample trajectories per day of training data}:
\figref{fig:RL:costJointEV}(a)
shows that
when $\Delta t \geq 3$ months, increasing the number of samples does not result in significant reduction in normalized cost of the learned policy (\ie $\CRL$) for both 10 and 50 charging stations.

The above analysis suggests that a training data time span of at least 3 months is needed to have a comparable performance over various number of samples per day and that when training data time span is at least 3 month long, smaller number of samples (of the order of 5K trajectories) can still achieve a comparable performance (with respect to training with larger samples per day). This answers \textbf{\qref{q:RL:training}}.

Next, we answer \textbf{\qref{q:RL:optimal}} (\ie how does the RL policy perform compare to an optimal all-knowing oracle algorithm?) by referring to the best performance measures in \figref{fig:RL:costJointEV}, for coordinating 10 and 50 EV charging stations.
We observe that the best performance is achieved when $\Delta t = 9$ months for both scenarios. The relative improvement in terms of reduction of normalized cost, compared to a business-as-usual uncontrolled charging scenario, $\Cbau$, amounts to 39\% and 30.4\% for 10 and 50 charging stations respectively.
Note that $\CRL$ is still 13\% and 15.6\% more expensive than the optimal policy cost $\Copt$ (the optimal policy would achieve 52\% reduction in cost with respect to $\Cbau$) for 10 and 50 charging stations respectively.
Still, it is important to realize that to find the optimal policy, we assume perfect knowledge of future EV charging sessions, including arrival and departure times and the energy requirements.
Clearly, having such complete knowledge of the future is not feasible in a real-world scenario: the proposed RL approach, which does not require such knowledge, thus is a more practical solution.

Finally, comparing the variance of the different runs (shaded regions in \figref{fig:RL:costJointEV} for 10 \vs 50 EV charging stations reveals that there is an increase in the variance between simulation runs when the group size is increased. 
Note that the same training horizons are used for both groups for a given $\Delta t$ and simulation run.
After observing the distributions of EV arrivals, EV departures and energy requirements, we conclude that high variability between the runs in \figref{fig:RL:costJointEV} does not stem from differences in the distributions among the various charging stations.
We rather hypothesize that this increased performance variance among runs is caused by the fact that the state-action space for coordinating the charging of 50 cars is considerably bigger than the one of 10 cars, given Eq.~\eqref{eq:RL:growth}.
The performance of the fitted Q-iteration is indeed greatly influenced by the training set $\mathcal{F}$ at the input of the algorithm.
With random sampling, there is no guarantee that most crucial parts of the state-action space (\eg best and worst trajectories) will be included in the training set $\mathcal{F}$. With larger trees, such a possibility is even more limited.
Re-exploration of the state-action space with a trained agent and retraining is one way to improve the performance. Efficient exploration of large state-action spaces is one of the active research domains in reinforcement learning and many algorithms are proposed to tackle the exploration problem (\eg \cite{Tang2017} and \cite{Mnih2016}). A summary of the exploration algorithms is presented in \cite{Mcfarlane2018}.
Such tackling of efficient exploration of the state-action space is left for future research.\footnote{As indicated previously, we limit this paper's focus to proposing the (scalable/generalizable) MDP formulation and experimentally exploring the resulting RL-based EV charging performance using realistic EV data.}


\subsection{Variance of performance over time (\qref{q:RL:time-variance})}
\label{sec:RL:results_time_variance}

\begin{figure}[!t]
	\centering
	\includegraphics[scale=0.8]{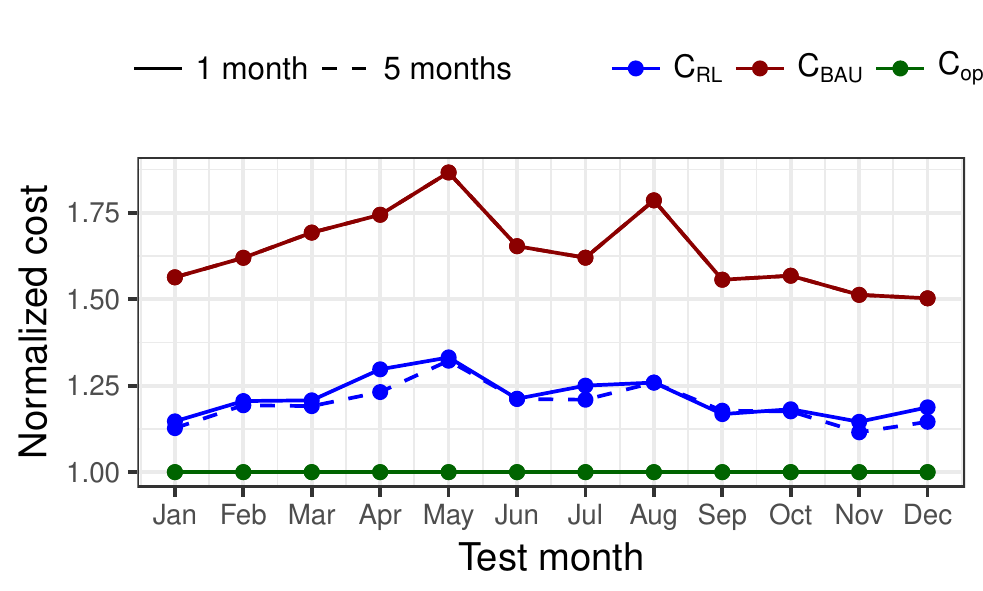}
	\caption{Performance using different months as test set and different time spans of the training set (1--5 months).}
	\label{fig:RL:Vartest}
\end{figure}

In the analyses presented in \figref{fig:RL:costJointEV}, the days in the last quarter of 2015 from the ElaadNL dataset were used to construct the test set.
Now, we investigate whether changing the test set influences the performance of the learned policy, as to answer the question how performance of our RL approach would vary over time throughout the year.
More specifically, we use each month of 2015 as a separate test set, using the preceding months as training data. We also vary the training data time span from 1 to 5 preceding months.
\Figref{fig:RL:Vartest} shows the normalized costs for coordinating $N_s=10$ charging stations. This is complemented in \figref{fig:RL:Vartest} with the relative cost improvement compared to the business-as-usual scenario, $\Cbau$.

\begin{figure}[!t]
	\centering
	\includegraphics[scale=0.75]{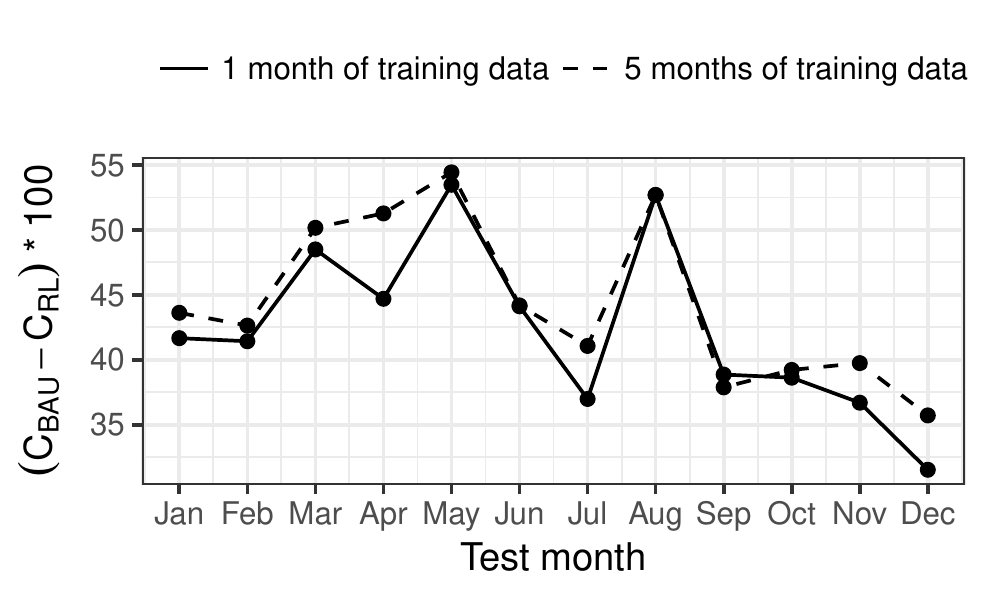}
	\caption{Improvement in normalized cost of the learned policy (RL) with respect to the business-as-usual policy (BAU).}
	\label{fig:RL:VartestBar}
\end{figure}  

\Figref{fig:RL:Vartest} shows that $\Cbau$ varies across the test months: for some months (\eg May and Aug), the difference $\Cbau-\Copt$ is larger than for others.
This indicates that the charging sessions in these test months have higher flexibility, which is exploited by the optimum solution.
For such months with higher $\Cbau-\Copt$, our proposed RL approach also achieves a higher reduction in normalized cost compared to $\Cbau$, as seen in \figref{fig:RL:VartestBar}.
Still, the achieved $\CRL$ is more expensive than $\Copt$ compared to the months who offer less flexibility.
We found that the days in which the optimal charging pattern requires the exploitation of larger charging delays are more challenging to learn by RL approach, in the sense that RL has greater difficulty in approaching the optimum (\ie obtaining higher $\CRL$).
One reason is the scarcity of such days in the training set, which results in imbalanced training data. Another reason is the random sampling of the large state action space, which does not guarantee inclusion of the scarce (but crucial) parts of the state-action space in the training set that is fed to the FQI algorithm.  

We further investigate the effect of increasing the training data time span from 1 preceding month to 5 preceding months for each test set. We find that for majority of the months, this results in improvement with respect to $\Cbau$ as depicted in \figref{fig:RL:VartestBar}. 

The analysis in this section reveals the following answer to \textbf{\qref{q:RL:time-variance}} (\ie How does the performance vary over time using realistic data?):
the RL algorithm performance depends on the available flexibility, with greater flexibility (expectedly) leading to larger cost reductions compared to the BAU uncontrolled charging, but greater difficulty in approaching the optimum performance.

\begin{figure}[!t]
	\centering
	\includegraphics[scale=0.45]{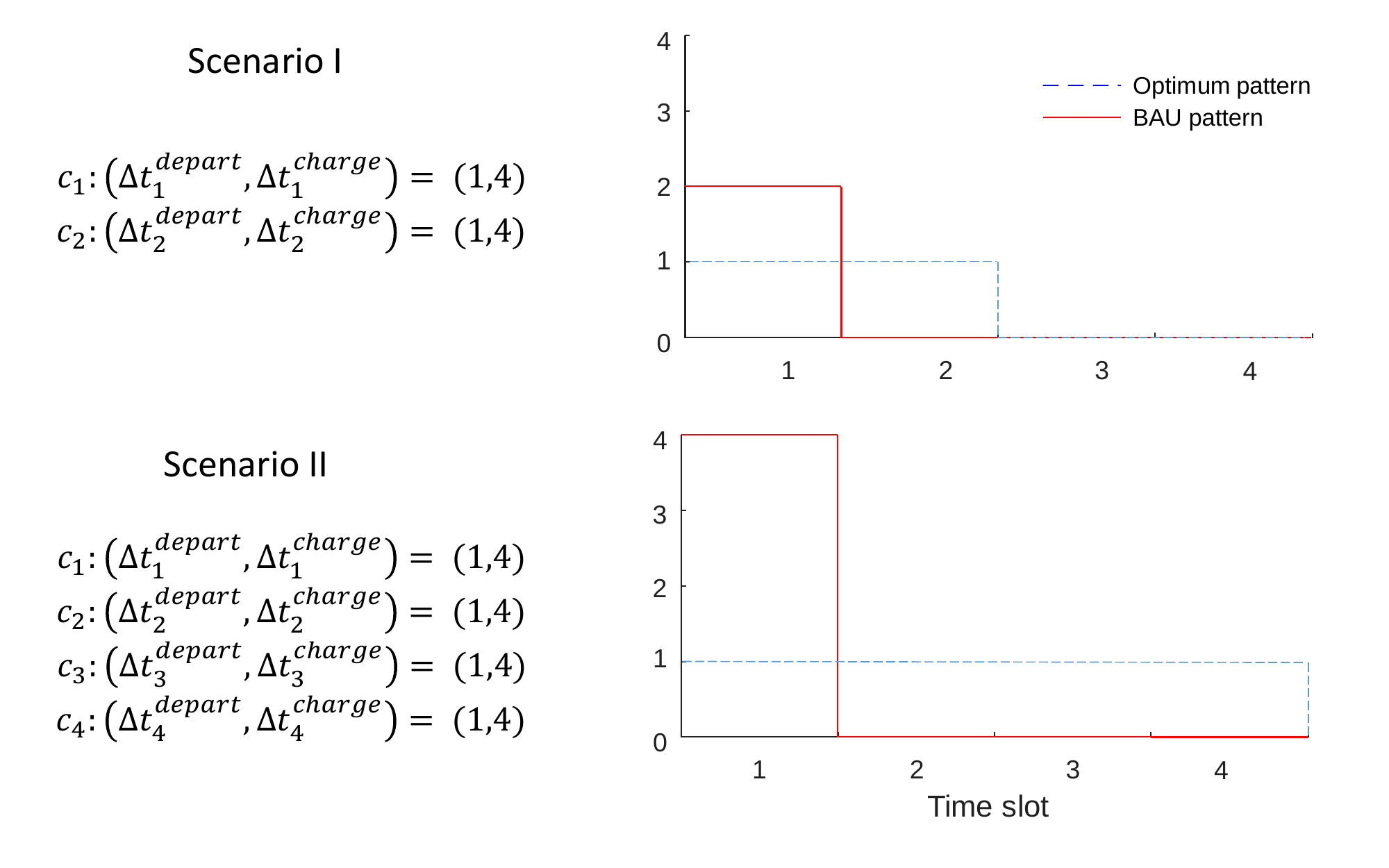}
	\caption{The effect of scaling up the group size on a normalized cost of a policy learned from 10 EV charging stations}
	\label{fig:RL:example}
\end{figure}  

\subsection{Generalization to Larger Scales (\qref{q:RL:generalization})}
\label{sec:RL:results_generalization}

While model-free approaches based on RL eliminate the need for accurate knowledge of the future EV session characteristics (as opposed to optimization based approaches), they still require a reasonably long training time to be able to efficiently coordinate the EV charging sessions. The runtime for the largest training set size (covering 9 months, with 5K sample trajectories per day) is approximately 3~hours for 10 EV charging stations, while that of 50 charging stations is approximately 48~hours.\footnote{Running on an Intel Xeon E5645 processor, 2.4\,GHz, 290\,GB RAM.} 

Since our proposed formulations are independent of the number of EV charging stations ($\Nmax$), it is interesting to investigate how a policy learned based on training with a small number of EV charging stations performs when applied to coordinate a larger group of stations.
To do this, we use the policy learned from data of 10 EV charging stations with $\Delta t=9$ months. We use the EV sessions in the last quarter of 2015 as our test set. To investigate the effect of the increase in the number of EV charging stations without changing other system characteristics, we duplicate the EV charging sessions by a factor \textit{scale} to create a test set of larger $\Nmax$.
This still changes the optimum solution as illustrated with a simple example in \figref{fig:RL:example} where the length of the control horizon is $\Smax = 4$ slots. In Scenario~I of \figref{fig:RL:example}, at time $t = 1$ we have 2 connecting cars: $V = \{(\tdepart_1,\tcharge_1) = (1,4), (\tdepart_2,\tcharge_2) = (1,4)\}$ and no other arrivals during the control horizon. The best action is to charge 50\% of the cars at $t = 1$ and $2$ to flatten the load curve.
In Scenario~II of \figref{fig:RL:example}, set $V$ is duplicated once and the best action now is to charge 25\% of the cars
in each of the control timeslots.

\begin{figure}[!t]
	\centering
	\includegraphics[scale=0.7]{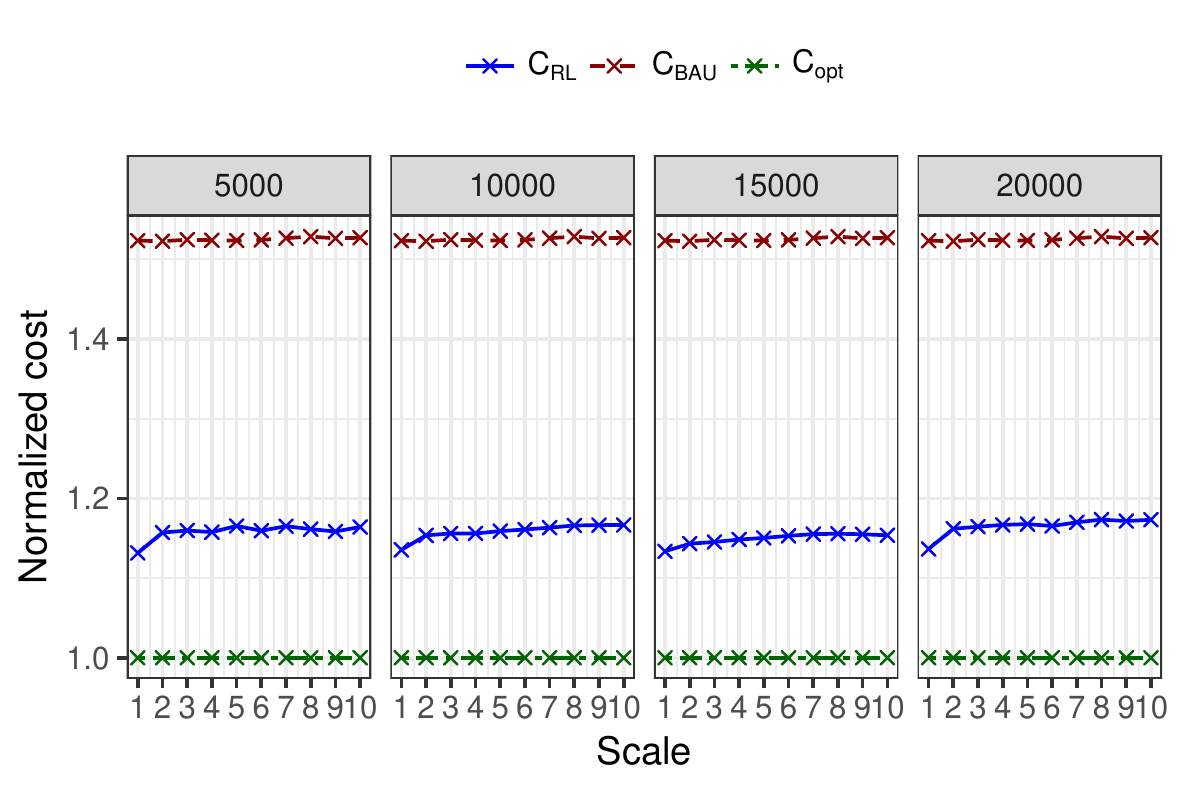}
	\caption{The effect of scaling up $\Nmax$ on a normalized cost of a policy learned from $\Nmax=10$ EV charging stations
    for different of number of sampled trajectories (ranging from 5K to 20K).}
	\label{fig:RL:scale}
\end{figure}  

The normalized costs (\ie relative to the optimum $\Copt$) of the learned policy for scaled-up group sizes are shown in \figref{fig:RL:scale} for various scales and number of samples per day in the training set. The scale of 1 corresponds to the original test set without any duplication.
\newtext{The largest jumps in normalized costs $\CRL$ are}
observed when the group size is doubled (\ie scale factor $2\times$). Further increases in $\Nmax$ (\ie more than $2\times$), only lead to marginal increase in normalized cost for any number of sample trajectories per day (ranging from 5K to 20K). These analyses further confirm that our proposed MDP formulations are generalizable to various group sizes and that a policy learned from a smaller group of EV charging stations can be used to coordinate the charging of a larger group, at least provided that the distribution of EV arrivals, departures and energy demands are similar. 

\section{Conclusion}
\label{sec:RL:conclusion}
In this paper, we took the first step to propose a reinforcement learning based approach for jointly controlling a charging demand of a group of EV charging stations. We formulated an MDP with scalable representation of an aggregated state of the group which effectively takes into account the individual EV charging characteristics (\ie arrival time, charging and connection duration). The proposed formulations are also independent of the number of charging stations and charging rates, hence, they generalize to varying number of charging stations. We used a real-world EV charging dataset to experimentally evaluate the performance of the proposed approach 
compared to an uncontrolled business-as-usual (BAU) policy, as well as an optimum solution that has a perfect knowledge of the EV charging session characteristics (in terms of arrival and departure times).
The summary of our analyses (in form of answer to 4 research questions) and the conclusions thereof, for a realistic 1-year long dataset (from ElaadNL)\cite{Nasrin2017}, are as follows:
    
\begin{enumerate}[label=(\arabic*)]
	\item While the representation of the state and action are independent of the group size (\ie number of charging stations), the resulting state-action space is still relatively large. Hence, feeding the entire state-action space to the learning algorithm (\ie FQI) is not feasible. 
    This raised the question \textbf{\qref{q:RL:training}}: What are appropriate training data time span and number of sampled trajectories from the decision trees? We investigated the effect of the training data time span and the number of sample trajectories per day on the performance of the learned policy and concluded that when the training data time span is longer than 3 months, a smaller number of samples (order of 5K) from each of the training days achieve similar performance as the larger number of sampled trajectories from those training days. 
	\item We investigated how the RL policy performs compared to an optimal all-knowing oracle algorithm (\ie \textbf{\qref{q:RL:optimal}}). We show that our proposed approach learns a policy which can reduce the normalized cost of coordinating charging across 10 and 50 EV charging stations by 39\% and 30.4\% respectively from the normalized cost of the uncontrolled BAU charging policy. The achieved reduction in performance by our approach does not require future knowledge about EV charging sessions and it is only 13\% (for $\Nmax=$10 charging stations) and 15.6\% (for $\Nmax=$50 charging stations) more expensive than the optimum solution cost with has a perfect knowledge of future EV charging demand. 
	\item We then analyzed how the performance of our  proposed RL approach varies over time using realistic data (\ie \textbf{\qref{q:RL:time-variance}}) by checking whether the learned policy performs similarly when various months of the year are used as test set while the agent is trained on the preceding months. The results indicate that the flexibility --- hence reduction in the normalized cost --- varies across various months. In particular, the months with larger flexibility have larger reduction in cost by the learned policy with respect to the normalized cost of the BAU policy.
Still, the cost gap between the learned policy and the optimal one is larger for those higher flexibility months. 
This is due to the scarcity of the days with larger flexibility in the training set as well as the random sampling of the state-action space, which does not guarantee inclusion of the rare but crucial parts of the state-action space in the training set that is fed to the FQI algorithm. 
	\item Finally, we trained an agent using an experience from 10 EV charging stations and applied the learned policy to control a 
higher number of charging stations (up to a factor of $10\times$ more arrivals) to check whether the learned approach generalizes to different group sizes (question \textbf{\qref{q:RL:generalization}}).
These analyses further confirmed that our proposed MDP formulations are generalizable to groups of varying sizes and that a policy learned from a small number of EV charging stations may be used to coordinate the charging of a larger group, at least provided that the distribution of EV arrivals, departures and energy demands are similar. 
\end{enumerate}

In our future research, we will study two three possible improvements to the presented approach:
\begin{enumerate}[label=(\arabic*)]
\item We used random exploration of state-action space to collect the experience (in form of tuples) as an input to our learning algorithm. We will investigate whether incorporating an efficient exploration strategies to perform a more informed sampling of the state-action space improves the performance? 
\item \newtext{We used a fully connected neural network for function approximation in the FQI algorithm. Since we represent our aggregate demand in a state using a matrix, it is relevant to investigate whether using convolutional neural networks (similar to the function approximation adopted in \cite{Claessens2018}) will further improve the performance.}
\item The learning algorithm in our proposed approach is based on the value iteration approach where the state-action value is estimated for various state-action pairs and an optimum policy is deduced form the estimated action-values. We will investigate whether learning the policy directly using policy iteration methods improves the performance.
\end{enumerate}


%

\section*{Acknowledgment}

The authors would like to thank Professor Pascal Poupart for his expert advise on the reinforcement learning algorithms, and Bert Claessens for providing constructive feedback on a draft of this paper.

\ifCLASSOPTIONcaptionsoff
  \newpage
\fi



\bibliographystyle{IEEEtran}
\bibliography{EV_RL_journal}

\end{document}